\documentclass[11pt]{article}

\usepackage[]{acl}

\usepackage{times}
\usepackage{latexsym}
\usepackage[T1]{fontenc}
\usepackage[utf8]{inputenc}
\usepackage{microtype}
\usepackage{inconsolata}
\usepackage{graphicx}
\usepackage{amsmath} 
\usepackage{amssymb}
\usepackage{booktabs}
\usepackage{array}
\usepackage{multirow}
\usepackage{colortbl}
\usepackage{enumitem}
\usepackage[most]{tcolorbox}
\usepackage{algorithm}
\usepackage{algorithmic}
\usepackage[normalem]{ulem}
\usepackage{xcolor}

\newcolumntype{L}[1]{>{\raggedright\arraybackslash}p{#1}}
\definecolor{rqyellow}{RGB}{255, 251, 226}
\definecolor{cvprblue}{rgb}{0.21,0.49,0.74}

\newtcolorbox{researchquestionboxenv}{
  colback=rqyellow,
  colframe=black,
  boxrule=0.55pt,
  arc=2mm,
  left=6pt,
  right=6pt,
  top=5pt,
  bottom=5pt,
  boxsep=0pt,
  before skip=8pt,
  after skip=8pt,
  enhanced
}

\newcommand{\method}{\textbf{\textsc{Tyler}}}

\title{\textcolor{cvprblue}{\method{}}: \textcolor{cvprblue}{Ty}ped \textcolor{cvprblue}{L}at\textcolor{cvprblue}{e}nt \textcolor{cvprblue}{R}easoning for Language Models ——\\When to Think, What to Compute, and How Much to Allocate}

\author{
Hanyu Lin$^{1}$ \quad
Min Cai$^{2}$ \quad
Jiawei Wen$^{1}$ \quad
Haodi Zhang$^{1}$ \\
$^{1}$Shenzhen University \quad
$^{2}$University of Alberta \\
\href{https://typed-latent-reasoning.github.io}{\texttt{typed-latent-reasoning.github.io}}
}

\begin{document}
\maketitle

\begin{abstract}
Chain-of-thought (CoT) prompting improves reasoning in large language models (LLMs) by externalizing intermediate computation as discrete text tokens, but this textual interface also introduces redundancy and inference overhead.
Latent reasoning offers a promising alternative by carrying part of the computation in continuous representations.
However, existing methods typically predefine when latent computation is invoked and how it is allocated during decoding, leaving a key problem unresolved: when to invoke latent computation, what type of computation to perform, and how much budget to allocate.
We propose \textbf{Ty}ped \textbf{L}at\textbf{e}nt \textbf{R}easoning (\method{}), a typed and budget-aware framework for latent reasoning during autoregressive decoding.
\method{} learns a policy that, at each decoding step, chooses between emitting a text token and switching to a latent computation module specialized for a particular reasoning function.
Once invoked, an operator maps the current reasoning state into latent tokens that support global planning, local state updates, or reusable procedural abstraction.
Across extensive experiments on three backbone LLMs, \method{} improves accuracy by up to 14.49 points over CoT and by up to 4.30 points over the strongest competing baseline.
It further generalizes across diverse reasoning domains and achieves the best final-stage performance with the lowest forgetting.
\end{abstract}

\section{Introduction}

Large Language models~(LLMs) typically solve complex reasoning tasks by generating explicit chain-of-thought~(CoT) token sequences~\citep{wei2022chain,wang2022selfconsistency,jaech2024openai,guo2025deepseek}.
Although CoT improves reasoning performance, it requires intermediate computation to be externalized as visible text.
This requirement increases redundancy generation and inference cost.
It also imposes a rigid interface: intermediate reasoning must be written out before the final answer is produced, rather than invoked internally on demand during autoregressive decoding.

Latent reasoning provides a promising way to relax this interface by keeping part of the intermediate computation in continuous representations, instead of unfolding the full reasoning process into explicit tokens~\citep{zhu2025survey,hao2024coconut,xu2025softcot,zhang2025softthinking}.
From this perspective, latent reasoning should not be viewed merely as a compressed form of CoT.
Rather, it can serve as a computation mechanism that allows the model to choose between visible decoding and silent latent computation during generation.

\begin{figure}[t]
\centering
\includegraphics[width=\columnwidth]{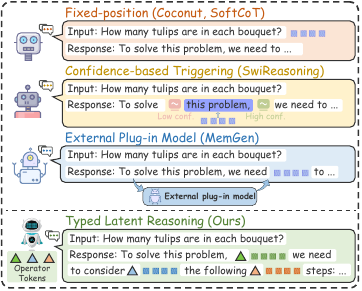}
\caption{
Comparison of latent reasoning paradigms.
Existing methods rely on fixed-position latent tokens, confidence-triggered latent computation, or external trigger models.
\method{} instead allows the LLM to interleave between visible decoding and typed latent-operator invocation during generation.
}
\label{fig:lcflow_comparison}
\end{figure}

Despite encouraging progress, two limitations remain.
First, prior work mainly focuses on \emph{how} to construct latent tokens, such as from hidden states~\citep{hao2024coconut,liu2026latentthoughts}, mixtures over token-embedding distributions~\citep{zhang2025softthinking,shi2025swireasoning}, or auxiliary soft-thought generators~\citep{xu2025softcot,xu2025softcotpp,zhang2025memgen}.
This construction-centered view often treats latent tokens as general-purpose carriers of intermediate computation.
However, explicit reasoning chains are functionally heterogeneous: different steps may orient the problem, update the evolving reasoning state, or reuse procedural solution patterns.
A latent reasoning mechanism should preserve these functional distinctions rather than collapse them into a single undifferentiated representation.

Second, existing methods remain limited in how they adaptively coordinate explicit reasoning and latent computation during generation.
Fixed-position methods~\citep{hao2024coconut,xu2025softcot} decouple latent computation from the evolving reasoning state, while adaptive methods mainly decide whether or where to enter latent space based on uncertainty signals~\citep{shi2025swireasoning,liu2026latentthoughts} or an external trigger model~\citep{zhang2025memgen}.
They do not allow the LLM itself to decide, during autoregressive decoding, which type of latent computation to perform or how much computation budget to allocate.
This gap motivates the following research question:

\begin{tcolorbox}[colframe=black!50, colback=blue!5, boxrule=1.5pt, arc=2mm, top=4pt, bottom=4pt, left=4pt, right=4pt, boxsep=1pt]
\raisebox{-0.2\baselineskip}{\includegraphics[height=1\baselineskip]{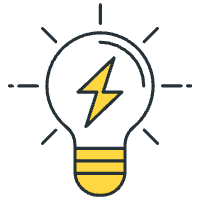}}
\textit{Can an LLM learn to interleave visible decoding with silent latent computation, while deciding when to think, what type of computation to perform, and how much budget to spend?}
\end{tcolorbox}

To answer this question, we propose \method{}, a typed latent reasoning framework that learns when to invoke latent computation, which type of computation to perform, and how much budget to allocate.
During generation, the LLM chooses between visible-token decoding and three typed latent operators: $O_g$ for global orientation, $O_s$ for state update, and $O_p$ for procedural abstraction.
Each operator maps the current reasoning state to a sequence of operator-conditioned latent tokens, which then guide subsequent visible decoding.
Because these operators are defined by reusable computational roles rather than task-specific output formats, \method{} supports latent computation that can transfer across reasoning domains.

The training procedure consists of two stages.
Stage~1 optimizes the latent operators and the Latent Synthesis Model while keeping the LLM frozen, aligning each operator with its intended reasoning function.
Stage~2 optimizes an operator-invocation policy with Group Relative Policy Optimization~(GRPO), updating the policy parameters while keeping the latent synthesis interface learned in Stage~1 fixed.
An operator-anchored auxiliary objective further stabilizes policy learning at sparse invocation positions.

% Empirically, \method{} achieves consistent improvements across multiple backbones and benchmarks, while demonstrating strong cross-domain generalization.
% Further analysis shows that the learned policy invokes different operators across task complexities and reasoning states, suggesting that these operators acquire differentiated functional roles.
% % Under sequential adaptation across code, science, math, and theorem reasoning, \method{} also achieves a stronger retention--adaptation trade-off.
% Under sequential adaptation across four domain, \method{} further achieves higher final performance with lower forgetting, indicating stronger continual adaptation.

% Our contributions are summarized as follows:
% \begin{itemize}
% \item We formulate latent reasoning as an online, typed, and budgeted computation problem, where an LLM learns when to invoke latent computation, which operation to perform, and how much budget to allocate.
% \item We introduce \method{}, a typed latent reasoning framework that interleaves visible decoding with latent operators for global orientation, state update, and procedural abstraction.
% \item We conduct extensive experiments across backbones and benchmarks, showing that \method{} generalizes beyond the training domain and improves accuracy over the strongest competing baseline by up to $4.30$ \%.
% Under sequential adaptation, \method{} further demonstrates strong continual-learning ability, achieving higher final performance while reducing forgetting.
% \end{itemize}
Empirically, \method{} achieves consistent improvements across multiple backbones and benchmarks, with gains that transfer beyond the training domain.
Further analysis shows that the learned policy invokes different operators across task difficulties and reasoning stages, suggesting that these operators acquire differentiated functional roles.
In a sequential adaptation setting spanning code, science, math, and theorem reasoning, \method{} obtains higher final performance with lower forgetting, indicating stronger continual adaptation.

Our contributions are summarized as follows:
\begin{itemize}
\item We formulate latent reasoning as an online, typed, and budgeted computation problem, where an LLM learns when to invoke latent computation, which operation to perform, and how much budget to allocate.
\item We introduce \method{}, a typed latent reasoning framework that interleaves visible decoding with latent operators for global orientation, state update, and procedural abstraction.
\item We conduct extensive experiments across multiple backbones and reasoning benchmarks, showing that \method{} improves accuracy over strong baselines, generalizes beyond the training domain, and exhibits stronger continual adaptation under sequential domain shifts.
\end{itemize}
\section{Related Work}

\paragraph{Explicit reasoning.}
CoT improves the reasoning ability of LLMs by making intermediate computation explicit as text~\citep{wei2022chain}. Subsequent work extends this paradigm through sampled reasoning paths~\citep{wang2022selfconsistency}, external actions and tool use~\citep{yao2022react,schick2023toolformer}, and search-based method~\citep{yao2023tree}. Reinforcement learning has also become central to scaling explicit reasoning behavior, ranging from PPO-based RLHF~\citep{schulman2017ppo,ouyang2022training} to recent methods based on GRPO-style optimization~\citep{shao2024deepseekmath,yu2025dapo}. These methods strengthen reasoning mainly by increase the computational load. 
In contrast, \method{} preserves the autoregressive decoding interface while allowing the model to adaptively choose between emitting visible tokens and performing silent latent computation.

\begin{figure*}[t]
  \centering
  \includegraphics[width=\textwidth]{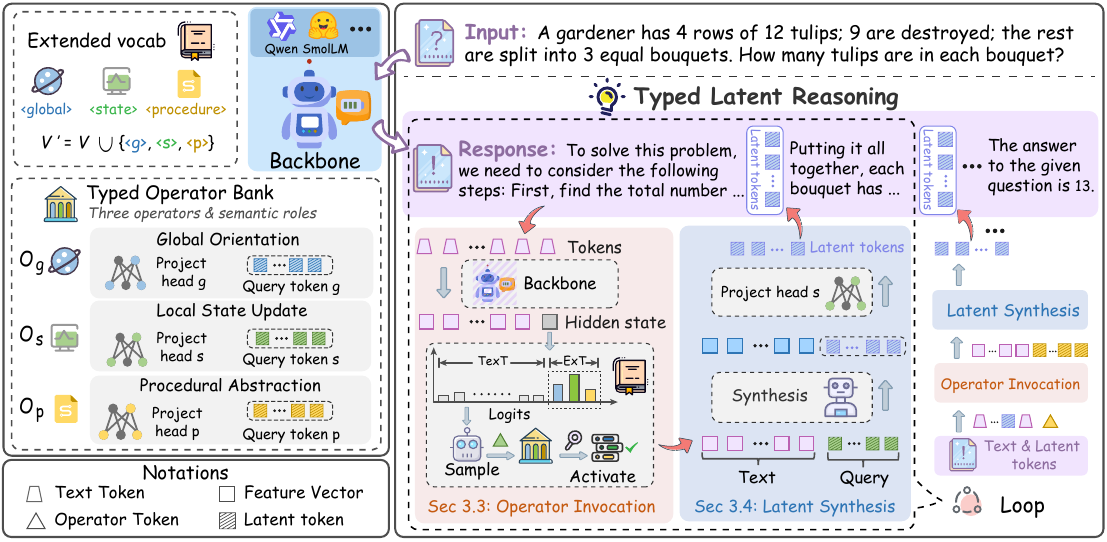}
  \caption{Overview of \method{}. During decoding, the backbone interleaves between visible text tokens and typed operators. When invoked, each operator maps the current context to an operator-conditioned sequence of latent tokens, which is inserted back into the context to guide subsequent decoding.
}
  \label{fig:lcf}
\end{figure*}

\paragraph{Latent reasoning.}
Latent reasoning replaces part of the visible tokens with continuous representations~\citep{zhu2025survey,chen2025reasoning}. Existing methods mainly study how latent tokens are constructed, including recycled hidden states~\citep{hao2024coconut,shen2025codi}, soft thought tokens~\citep{xu2025softcot,wei2025sim}, and embedding-space mixtures~\citep{zhang2025softthinking}. Other work studies when to activate latent computation through confidence signals~\citep{shi2025swireasoning,liu2026latentthoughts}, enternal trigger model~\citep{zhang2025memgen}, or pause tokens~\citep{goyal2023pause,pfau2024dot}.
These methods demonstrate the value of latent reasoning, but they usually treat latent computation as a uniform form.
They provide limited study for deciding what functional role latent computation should serve or how much computation budget should be allocated. 
In contrast, \method{} formulates latent reasoning as an online, typed, and budgeted computation problem, where latent operators serve different computational roles and are invoked under an explicit budget.
\section{Methodology}
\label{sec:methodology}

\subsection{Preliminaries}
\label{sec:method-prelim}

We consider a decoder-only autoregressive language model parameterized by
$\theta$, which defines a next-token distribution
$p_{\theta}(\cdot \mid x_t)$ over a discrete vocabulary $V$ given an input
embedding sequence $x_t$. Let $d$ denote the embedding dimension, and let
$x_t\in\mathbb{R}^{T_t\times d}$ be the decoding context at step $t$, where
$T_t$ is the current sequence length. In standard autoregressive decoding, the
model samples a visible token
\begin{equation}
a_t \sim p_{\theta}(\cdot \mid x_t),
\label{eq:next-token-sampling}
\end{equation}
and appends its embedding to form the next-step input:
\begin{equation}
x_{t+1}=[\,x_t;\,\mathrm{emb}(a_t)\,].
\label{eq:ar-update}
\end{equation}

This scheme tightly couples intermediate computation with discrete token
generation: each reasoning state must be externalized through a
vocabulary-indexed embedding before it can influence subsequent decoding.
Latent reasoning addresses this constraint by allowing continuous vectors to be
inserted into the decoding context. Given the current embedding sequence $x_t$,
a set of latent tokens $z_t\in\mathbb{R}^{N_t\times d}$ can be appended as
\begin{equation}
x_{t+1}=[\,x_t;\,z_t\,].
\label{eq:latent-update}
\end{equation}
These latent tokens are consumed by the autoregressive model as internal computational states while remaining invisible to the user.

\subsection{Overview}
\label{sec:method-overview}
% \method{} extends autoregressive decoding with adaptive latent computation.
% The framework consists of two main components.
% First, the operator invocation policy decides whether the model should emit a visible token or invoke a typed operator (Section~\ref{sec:method-operators}).
% Second, the latent synthesis module maps the activated operator and the current context to a sequence of latent tokens (Section~\ref{sec:method-synthesis}).
% Figure~\ref{fig:lcf} illustrates the overall pipeline.
As shown in Figure~\ref{fig:lcf}, method{} extends autoregressive decoding with adaptive latent computation. The framework consists of two main components. First, the operator invocation policy decides whether the model should emit a visible token or invoke a typed operator (Section~\ref{sec:method-operators}). Second, the latent synthesis module maps the activated operator and the current context to a sequence of latent tokens (Section~\ref{sec:method-synthesis}).

\subsection{Operator Invocation}
\label{sec:method-operators}

We first define the latent operators and describe how \method{} invokes them
during autoregressive decoding. In explicit CoT reasoning, intermediate tokens
often play different functional roles: some provide global orientation, some
update the local reasoning state, and others express reusable procedural
patterns. This suggests that latent computation should not be treated as a
single undifferentiated operation.

Motivated by this observation, \method{} introduces three typed latent
operators. Each operator $O_k$ has learnable parameters $\omega_k$ and maps the
current decoding context to a sequence of latent tokens that supports
subsequent generation:
\begin{equation}
O_k(\cdot;\omega_k): x_t \mapsto z_t^k \in \mathbb{R}^{N_k \times d}.
\end{equation}
The operator type provides an inductive bias about the intended computational
role, while its behavior is learned from CoT data.

We instantiate three operators:
\begin{equation}
\mathcal{O} = \{O_g, O_s, O_p\},
\end{equation}
where $O_g$ supports global orientation and planning, $O_s$ supports local
state updates, and $O_p$ captures reusable procedural abstractions. These
operators are encouraged by heuristic position supervision in Stage~1
(Section~\ref{sec:stage1}) and regulated by a budget-aware routing policy in
Stage~2 (Section~\ref{sec:training}).

To enable operator invocation during decoding, we introduce one special token
for each operator. Let $\mathcal{U}=\{u_g,u_s,u_p\}$ denote the set of operator
tokens, where $u_k$ is associated with $O_k$. The decoding action space is
extended to $V' = V \cup \mathcal{U}$ by adding three trainable operator-token
rows $\psi \in \mathbb{R}^{3 \times d}$ to the LM head. The
next-token distribution is denoted as $p_{\theta,\psi}(\cdot\mid x_t)$, and
the model samples
\begin{equation}
a_t \sim p_{\theta,\psi}(\cdot\mid x_t), \qquad a_t\in V'.
\end{equation}

If $a_t$ is a visible token, the model appends its embedding and emits it to
the user. If $a_t=u_k$ is an operator token, the corresponding operator
synthesizes latent tokens $z_t^k=O_k(x_t;\eta_k)$, which are appended to the
context $x_t$:
\begin{equation}
x_{t+1}=
\begin{cases}
[\,x_t;\, \mathrm{emb}(a_t)\,], & a_t\in V,\\[3pt]
[\,x_t;\, z_t^k\,], & a_t=u_k\in\mathcal{U}.
\end{cases}
\label{eq:lcf-transition}
\end{equation}

The same policy is then applied to the updated context $x_{t+1}$. Thus,
visible-token emission and latent-operator invocation compete within a single
next-action distribution, without requiring an external controller.

\subsection{Latent Synthesis}
\label{sec:method-synthesis}

Once the LLM samples an operator $O_k$, \method{} must synthesize latent tokens
that are compatible with the LLM input-embedding space. Rather than directly
reusing hidden states~\citep{hao2024coconut} or mixing vocabulary-space
embeddings~\citep{zhang2025softthinking}, we follow MemGen~\citep{zhang2025memgen}
and add trainable LoRA layers~\citep{hu2021lora} to the backbone while keeping
the original parameters $\theta$ frozen. Thus, latent tokens are synthesized
through parameter-efficient adaptation without updating the backbone directly.

When $O_k$ is activated at step $t$, it uses the current context $x_t$ to
produce $N_k$ latent tokens. All operators share the LoRA-augmented synthesizer
$\mathrm{Synth}_{\theta,\phi}$, but specialize through operator-specific query
tokens $Q_k\in\mathbb{R}^{N_k\times d}$ and projection heads
$\mathrm{Proj}_k:\mathbb{R}^{d}\!\to\!\mathbb{R}^{d}$.

The activated operator appends its query tokens to the current context and performs a single forward pass through the LoRA-augmented LLM.
Because the synthesizer is decoder-only and the query tokens are placed after the context, the query positions can attend to the preceding context:
\begin{equation}
h_t^k =
\mathrm{Synth}_{\theta,\phi}\!\left([\,x_t;\, Q_k\,]\right)_{\mathrm{query}}
\in \mathbb{R}^{N_k \times d}.
\label{eq:synth-intermediate}
\end{equation}
The operator-specific projection head then maps the query-position hidden states into the LLM input-embedding space:
\begin{equation}
z_t^k = \mathrm{Proj}_k(h_t^k) \in \mathbb{R}^{N_k\times d}.
\label{eq:synth-final}
\end{equation}
Although all operators share $\mathrm{Synth}_\phi$ and read the same context $x_t$, they specialize through distinct query tokens $Q_k$, distinct projection heads $\mathrm{Proj}_k$, and context-aware supervision over invocation positions introduced in Section~\ref{sec:stage1}.

\section{Training}
\label{sec:training}

\method{} is trained in two stages.
Stage~1 learns to synthesize operator-conditioned latent tokens while keeping the backbone LLM fixed.
Stage~2 freezes the latent synthesis-specific parameters learned in Stage~1 and trains the LLM to invoke typed operators adaptively under a computation budget.
Complete training details are provided in Appendix~\ref{app:training-details}.

\subsection{Latent Synthesis Optimization}
\label{sec:stage1}

Stage~1 optimizes the LoRA parameters $\phi$ in the latent synthesizer
$\mathrm{Synth}_{\theta,\phi}$ and the operator-specific components
$\{Q_k,\mathrm{Proj}_k\}_{O_k\in\mathcal{O}}$, while keeping the backbone
LLM parameters $\theta$ frozen. We denote the trainable Stage~1 parameters by
$\Phi=\{\phi,\{Q_k,\mathrm{Proj}_k\}_{O_k\in\mathcal{O}}\}$.
For each supervised reasoning trace $(q,y_{1:T})$, we construct a set of
candidate boundary--operator pairs $\mathcal{B}(q,y)$ using structure-aware
heuristics. Specifically, $O_g$ is assigned to answer-onset positions, $O_s$
to reasoning-step boundaries, and $O_p$ to structural boundaries such as
formulas, code blocks, and structured fields. Details are provided in Appendix~\ref{app:training-details}.

Given a pair $(b,k)\sim\mathcal{B}(q,y)$, the selected operator
synthesizes latent tokens $z_b^k=O_k(x_b)$ from the prefix
$x_b=(q,y_{<b})$.
We insert $z_b^k$ into the teacher-forced trajectory, perform a single causal
forward pass, and compute the next-token cross-entropy only at visible
target-token positions:
\begin{equation}
\mathcal{L}_{\mathrm{stage1}}
=
-\mathbb{E}_{\mathcal{D},\mathcal{B}}
\sum_{t=1}^{T}
\log p_{\theta,\Phi}
\!\left(y_t \mid \tilde{c}_{b,t}^{k}\right).
\label{eq:operator-sft}
\end{equation}
Here, $\mathbb{E}_{\mathcal{D},\mathcal{B}}$ denotes the expectation over
$(q,y)\sim\mathcal{D}$ and $(b,k)\sim\mathcal{B}(q,y)$. $\tilde{c}_{b,t}^{k}$ denotes the latent-augmented context before predicting the visible token $y_t$. For $t<b$, this context reduces to the visible prefix $(q,y_{<t})$; for $t\ge b$, it additionally contains
the inserted latent tokens, i.e.,
$(q,y_{<b},z_b^k,y_{b:t-1})$. The inserted latent tokens are used only as
context and are not prediction targets. Gradients are applied only to
$\Phi$, while the backbone parameters $\theta$ remain frozen. Consequently,
Stage~1 trains each operator through visible next-token supervision while
preserving the backbone reasoner.

\subsection{Operator Invocation Optimization}
\label{sec:stage2}

Stage~2 learns a budget-aware operator-invocation policy that decides whether to emit a visible token or invoke latent computation, which typed operator to select, and how much budget to spend.
We freeze the Stage~1 synthesis parameters $\Phi=\{\phi,\{Q_k,\mathrm{Proj}_k\}_{O_k\in\mathcal{O}}\}$, including the latent synthesizer LoRA parameters and all operator-specific components.
The backbone base weights $\theta$ are also kept frozen. The decoding policy is adapted through a separate set of LoRA parameters $\eta$, together with the operator-token head rows $\psi$.
We denote the resulting policy distribution by $p_{\theta,\eta,\psi}$ and optimize $\{\eta,\psi\}$ with GRPO~\citep{shao2024deepseekmath}.
For each prompt $q$, the policy samples a group of trajectories. Each trajectory
is assigned a reward that combines task performance with a success-gated budget
penalty $P_{bud}$:
\begin{equation}
\begin{aligned}
P_{\mathrm{bud}}(\tau)
&=
s(\tau)\bigl(n_{\mathrm{op}}(\tau)-B_O\bigr)_+, \\
R(\tau)
&=
R_{\mathrm{task}}(\tau)
-
\lambda P_{\mathrm{bud}}(\tau).
\end{aligned}
\label{eq:trajectory-reward}
\end{equation}
Here, $s(\tau)$ indicates whether the trajectory successfully solves the task,
$n_{\mathrm{op}}(\tau)$ is the number of operator invocations, and $B_O$ is the invocation budget. The penalty is applied only to successful
over-budget trajectories, encouraging the policy to use latent computation
efficiently without discouraging exploration on failed trajectories.

Because operator invocations occupy only a small fraction of each generated
trajectory, sequence-level rewards provide weak credit-assignment signals for
operator selection. We therefore introduce an operator-anchored auxiliary
objective that applies the GRPO surrogate at operator-invocation
positions. The Stage~2 objective is
\begin{equation}
\mathcal{L}_{\mathrm{stage2}}
=
\mathcal{L}_{\mathrm{GRPO}}(R)
+
\alpha \mathcal{L}_{\mathrm{anch}},
\label{eq:stage2-objective}
\end{equation}
where $\mathcal{L}_{\mathrm{anch}}$ stabilizes optimization at sparse invocation
positions, and $\alpha$ controls the strength of this auxiliary objective.

\begin{table*}[!t]
  \centering
  \small
  \setlength{\tabcolsep}{4pt}
  \begin{tabular}{l<{\hspace{16pt}}lcccc<{\hspace{10pt}}c}
    \toprule
    \textbf{Backbone} & \textbf{Method} & \textbf{GSM8K} & \textbf{MATH-500} & \textbf{GPQA-Diamond} & \textbf{TheoremQA} & \textbf{Average} \\
    \midrule
    \multirow{9}{*}{\textbf{SmolLM3-3B}}
      & CoT                 & 58.91 & 68.00 & 19.70 & 19.54 & 41.54 \textcolor{black!60}{\scriptsize($\uparrow$0.00)} \\
      & SFT                 & 83.02 & 66.60 & 21.72 & 25.30 & 49.16 \textcolor{green!60!black}{\scriptsize($\uparrow$7.62)} \\
      & GRPO                & 72.71 & \underline{74.20} & 28.28 & 26.51 & 50.43 \textcolor{green!60!black}{\scriptsize($\uparrow$8.89)} \\
      \cmidrule(lr){2-7}
      & SoftCoT             & 83.85 & 69.20 & 24.24 & 25.03 & 50.58 \textcolor{green!60!black}{\scriptsize($\uparrow$9.04)} \\
      & Soft-Thinking       & 76.65 & 68.40 & 21.72 & 21.15 & 46.98 \textcolor{green!60!black}{\scriptsize($\uparrow$5.44)} \\
      & MemGen              & \underline{84.15} & 70.20 & 25.25 & 27.30 & 51.73 \textcolor{green!60!black}{\scriptsize($\uparrow$10.19)} \\
      & SwiReasoning        & 74.98 & 68.60 & 20.70 & 19.40 & 45.92 \textcolor{green!60!black}{\scriptsize($\uparrow$4.38)} \\
      \cmidrule(lr){2-7}
      \rowcolor{blue!6}
      \cellcolor{white} & \textbf{\method{} } \textcolor{black!62}{\scriptsize Stage~1} & 83.24 & 71.20 & \underline{30.50} & \underline{28.76} & \underline{53.43} \textcolor{green!60!black}{\scriptsize($\uparrow$11.89)} \\
      \rowcolor{blue!12}
      \cellcolor{white} & \textbf{\method{} } \textcolor{black!62}{\scriptsize Stage~2} & \textbf{85.21} & \textbf{76.40} & \textbf{33.50} & \textbf{29.00} & \textbf{56.03} \textcolor{green!60!black}{\scriptsize($\uparrow$14.49)} \\
    \midrule
    \multirow{9}{*}{\textbf{Qwen3-4B}}
      & CoT                 & 90.52 & 82.60 & 16.16 & 36.95 & 56.56 \textcolor{black!60}{\scriptsize($\uparrow$0.00)} \\
      & SFT                 & 86.96 & 72.80 & 21.72 & 35.61 & 54.27 \textcolor{red!70!black}{\scriptsize($\downarrow$2.29)} \\
      & GRPO                & 90.14 & 81.20 & 39.90 & \underline{39.36} & 62.65 \textcolor{green!60!black}{\scriptsize($\uparrow$6.09)} \\
      \cmidrule(lr){2-7}
      & SoftCoT             & 88.32 & 72.60 & 38.38 & 34.94 & 58.56 \textcolor{green!60!black}{\scriptsize($\uparrow$2.00)} \\
      & Soft-Thinking       & 91.51 & 81.40 & 15.15 & 36.41 & 56.12 \textcolor{red!70!black}{\scriptsize($\downarrow$0.44)} \\
      & MemGen              & 89.65 & 78.20 & \underline{40.40} & 36.87 & 61.28 \textcolor{green!60!black}{\scriptsize($\uparrow$4.72)} \\
      & SwiReasoning        & 90.21 & 82.00 & 17.17 & 33.50 & 55.72 \textcolor{red!70!black}{\scriptsize($\downarrow$0.84)} \\
      \cmidrule(lr){2-7}
      \rowcolor{blue!6}
      \cellcolor{white} & \textbf{\method{}} \textcolor{black!62}{\scriptsize Stage~1} & \underline{91.66} & \underline{83.00} & 39.90 & 38.95 & \underline{63.38} \textcolor{green!60!black}{\scriptsize($\uparrow$6.82)} \\
      \rowcolor{blue!12}
      \cellcolor{white} & \textbf{\method{}} \textcolor{black!62}{\scriptsize Stage~2} & \textbf{92.87} & \textbf{85.80} & \textbf{43.43} & \textbf{41.02} & \textbf{65.78} \textcolor{green!60!black}{\scriptsize($\uparrow$9.22)} \\
    \bottomrule
  \end{tabular}
  \caption{Results on \textbf{SmolLM3-3B} and \textbf{Qwen3-4B}. All values are Pass@1 accuracy (\%). We highlight the \textbf{best} and \underline{second-best} results. Improvements across both backbones indicate that \method{} generalizes well across multiple task domains.}
  \label{tab:main}
\end{table*}

\section{Experiments}
\label{sec:experiments}

We organize the evaluation around five questions: whether \method{} (i) generalizes across domains, (ii) retains continual-learning capabilities during sequential task adaptation, (iii) benefits from each proposed component, (iv) invokes operators in a task-adaptive manner under a computation budget, and (v) induces distinguishable functional roles among the three typed operators.
% We study these questions in the main results (\S~\ref{sec:exp-main-results}), ablation studies (\S~\ref{sec:exp-ablation}), and operator analysis (\S~\ref{sec:exp-operator-analysis}).

\subsection{Experimental Setup}
\label{sec:exp-setup}

\paragraph{Backbones and training.}
We instantiate \method{} on Qwen2.5-1.5B-Instruct \citep{yang2024qwen25}, SmolLM3-3B \citep{bakouch2025smollm3}, and Qwen3-4B \citep{yang2025qwen3}, covering three model families in the 1.5B--4B range. Stage~1 and Stage~2 are trained on 15K problem--solution traces sampled from OpenR1-Math \citep{openr1}; the same data and prompt templates are used for all training-based baselines.

\paragraph{Benchmarks.}
We evaluate on four benchmarks covering three reasoning regimes: mathematical reasoning with GSM8K \citep{cobbe2021gsm8k} and MATH-500 \citep{hendrycks2021math}; scientific reasoning with GPQA-Diamond \citep{rein2023gpqa}; and theorem-oriented reasoning with TheoremQA \citep{chen2023theoremqa}.

\paragraph{Baselines.}
We compare \method{} against representative methods in two regimes. (i) {Explicit reasoning}: CoT \citep{wei2022chain}, SFT, and GRPO \citep{shao2024deepseekmath}. (ii) {Latent reasoning}: SoftCoT \citep{xu2025softcot}, Soft-Thinking \citep{zhang2025softthinking}, MemGen \citep{zhang2025memgen}, and SwiReasoning \citep{shi2025swireasoning}.

\paragraph{Metrics and implementation.}
We report Pass@1 accuracy under greedy decoding, and compute the average as an unweighted macro-average across the four benchmarks. Unless otherwise specified, Stage~1 uses latent-token lengths $(N_g,N_s,N_p)=(8,4,4)$, and Stage~2 uses a latent-call budget $B_O=5$. Implementation details are provided in Appendix~\ref{app:experiment-details}.

\begin{table*}[!t]
  \centering
  \footnotesize
  \setlength{\tabcolsep}{3pt}
  \begin{tabular}{llcccc!{\color{black!36}\vrule width 1pt}ccccc}
    \toprule
    & \textbf{Variant} & \textbf{Routing} & \textbf{Typed} & \textbf{Budget} & $\mathcal{L}_{\mathrm{anch}}$ & \textbf{GSM8K} & \textbf{MATH} & \textbf{GPQA} & \textbf{Theorem} & \textbf{Average} \\
    \midrule
    \emph{Per-stage}
    & (a) CoT             & --   & --  & --  & -- & 90.52 & 82.60 & 16.16 & 36.95 & 56.56 \textcolor{black!60}{\scriptsize($\uparrow$0.00)}  \\
    & (b) +Stage 1        & $\times$ & $\checkmark$ & --  & -- & 91.66 & 83.00 & \underline{39.90} & 38.95 & 63.38 \textcolor{green!60!black}{\scriptsize($\uparrow$6.82)} \\
    & (c) +Stage 2 (task-only) & $\checkmark$ & $\checkmark$ & $\times$ & $\times$ & \textbf{93.25} & 83.60 & 38.38 & 40.05 & \underline{63.82} \textcolor{green!60!black}{\scriptsize($\uparrow$7.26)} \\
    & (d) +Stage 2 (+budget) & $\checkmark$ & $\checkmark$ & $\checkmark$ & $\times$ & 91.57 & \underline{83.80} & 38.89 & \underline{40.42} & 63.67 \textcolor{green!60!black}{\scriptsize($\uparrow$7.11)} \\
    \rowcolor{blue!12}
    \cellcolor{white} & (e) \method{}    & $\checkmark$ & $\checkmark$ & $\checkmark$ & $\checkmark$ & \underline{92.87} & \textbf{85.80} & \textbf{43.43} & \textbf{41.02} & \textbf{65.78} \textcolor{green!60!black}{\scriptsize($\uparrow$9.22)} \\
    \midrule
    \emph{Typed?}
    & (f) Shared operators      & $\checkmark$ & $\times$ & $\checkmark$ & $\checkmark$ & 91.10 & 83.20 & 37.88 & 39.62 & 62.95 \textcolor{green!60!black}{\scriptsize($\uparrow$6.39)} \\
    & (g) Swap $O_s{\leftrightarrow}O_g$ & $\checkmark$ & swap & $\checkmark$ & $\checkmark$ & 92.49 & 80.40 & 35.35 & 33.50 & 60.44 \textcolor{green!60!black}{\scriptsize($\uparrow$3.88)} \\
    \midrule
    \emph{Routing?}
    & (h) Rand@25\%       & random  & $\checkmark$ & $\approx$ & -- & 90.71 & 82.80 & 34.85 & 38.81 & 61.79 \textcolor{green!60!black}{\scriptsize($\uparrow$5.23)} \\
    & (i) Rand@50\%       & random  & $\checkmark$ & $\approx$ & -- & 91.01 & 83.00 & 35.86 & 39.22 & 62.27 \textcolor{green!60!black}{\scriptsize($\uparrow$5.71)} \\
    & (j) Rand@75\%       & random  & $\checkmark$ & $\approx$ & -- & 91.15 & 82.60 & 36.36 & 38.65 & 62.19 \textcolor{green!60!black}{\scriptsize($\uparrow$5.63)} \\
    & (k) Rand@100\%      & random  & $\checkmark$ & --        & -- & 91.66 & 83.00 & \underline{39.90} & 38.95 & 63.38 \textcolor{green!60!black}{\scriptsize($\uparrow$6.82)} \\
    \bottomrule
  \end{tabular}
  \caption{Unified ablation on Qwen3-4B. All values are Pass@1 accuracy (\%);}
  \label{tab:unified-ablation}
\end{table*}

\subsection{Main Results}
\label{sec:exp-main-results}

\paragraph{Consistent Gains, Cross-Domain Generalization, and Robustness.}
\method{} achieves the best average performance on both backbones. On SmolLM3-3B, Stage~2 reaches an average accuracy of $56.03$, outperforming CoT by $14.49$ points and the strongest baseline, MemGen, by $4.30$ points. On Qwen3-4B, Stage~2 reaches $65.78$, improving over CoT by $9.22$ points and over the strongest prior baseline, GRPO, by $3.13$ points. Although \method{} is trained only on a subset of OpenR1-Math, its gains extend beyond mathematical reasoning to scientific and theorem-oriented benchmarks: Stage~2 improves GPQA-Diamond over CoT by $13.80$ and $27.27$ points on SmolLM3-3B and Qwen3-4B, respectively, and improves TheoremQA by $9.46$ and $4.07$ points. These results suggest that the learned latent computation transfers beyond the training domain. Existing baselines are more sensitive to the choice of backbone: SFT improves SmolLM3-3B but slightly degrades Qwen3-4B, while GRPO is stronger on Qwen3-4B than on SmolLM3-3B. Latent-reasoning baselines show similar variance, with MemGen being strongest on SmolLM3-3B but not on Qwen3-4B. In contrast, \method{} consistently achieves the best average performance and maintains gains on out-of-domain benchmarks, indicating stronger robustness across backbone choices and evaluation domains.

\begin{figure}[t]
  \centering
  \includegraphics[width=\columnwidth]{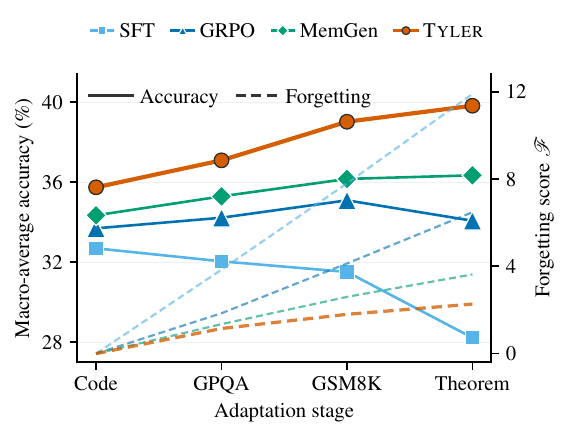}
  \caption{Continual-learning results on Qwen2.5-1.5B-Instruct. \emph{(a)} Macro-average after each adaptation stage. \emph{(b)} Forgetting score $\mathcal{F}$; lower is better.}
  \label{fig:continual-summary}
\end{figure}

\paragraph{Continual Learning.}
Figure~\ref{fig:continual-summary} evaluates \method{} under sequential adaptation across code, science, math, and theorem-oriented tasks. After the final adaptation stage, \method{} achieves the highest macro-average accuracy, reaching $39.8\%$ and outperforming the strongest baseline by about $3.5$ points. It also obtains the lowest forgetting score, with $\mathcal{F}=2.3$, reducing forgetting by about $1.4$ points compared with MemGen.
These results indicate that \method{} not only adapts more effectively to new reasoning domains, but also better preserves previously acquired capabilities. Additional details are provided in Appendix~\ref{app:continual-learning}.

\begin{figure*}[t]
  \centering
  \includegraphics[width=\textwidth]{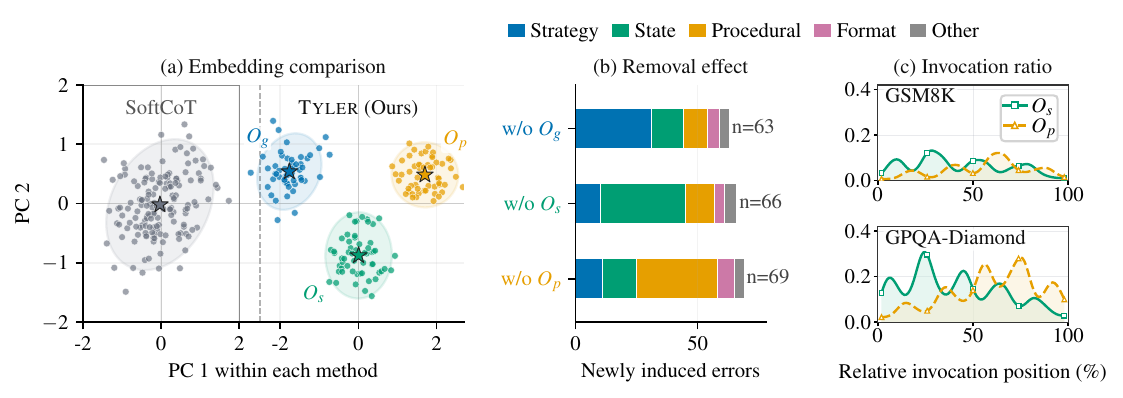}
  \caption{Diagnostics of typed latent operators with Qwen3-4B.
{(a)} PCA visualization of latent representations under the same embedding protocol.
{(b)} Error decomposition of newly induced failures after removing each typed operator on MATH-500.
{(c)} Relative invocation-position distributions on GSM8K and GPQA.}
  \label{fig:pca-typed}
\end{figure*}

\subsection{Ablation Studies}
\label{sec:exp-ablation}

The full \method{} combines operator-conditioned latent token synthesis, budget-aware operator invocation, and the operator-anchored loss $\mathcal{L}_{\mathrm{anch}}$. Table~\ref{tab:unified-ablation} ablates these components on Qwen3-4B by incrementally enabling Stage~1 and Stage~2, and by replacing typed operators or the learned router under comparable invocation settings. The Swap variant keeps the router, budget, and anchor loss fixed, but exchanges the synthesis paths of $O_s$ and $O_g$ at inference time.

Rows~(a)--(e) show that Stage~1 provides the largest gain, while Stage~2 improves selective operator use. Invoking learned typed operators at all candidate boundaries raises the average accuracy from $56.56$ to $63.38$, with GPQA-Diamond improving from $16.16$ to $39.90$, indicating transfer beyond in-domain math traces. Learned routing and budget control maintain this gain, and adding $\mathcal{L}_{\mathrm{anch}}$ further improves the average to $65.78$, outperforming CoT by $9.22$ points and Stage~1 by $2.40$ points. This suggests that anchored credit assignment is useful for sparse operator-invocation decisions.

Rows~(f)--(g) further show that the improvement is not merely due to latent capacity or access to candidate boundaries. Sharing operator-specific modules reduces the average to $62.95$, while swapping $O_s$ and $O_g$ drops it to $60.44$, indicating that the learned state-update and global-orientation operators play distinct roles. Rows~(h)--(l) show that random or fixed invocation schedules remain inferior. Overall, these results indicate that \method{} must jointly learn \emph{which} latent operator to invoke and \emph{where} to invoke it.

\begin{figure}[t]
  \centering
  \includegraphics[width=\columnwidth]{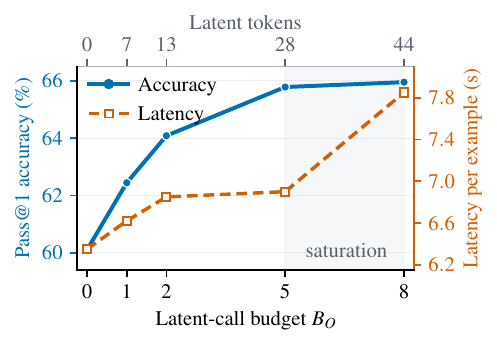}
  \caption{Latent-budget scaling on Qwen3-4B, averaged across GSM8K, MATH, GPQA, and TheoremQA.}
  \label{fig:latent-budget-scaling}
\end{figure}

\subsection{Additional Analysis}
\label{sec:exp-operator-analysis}

We further analyze whether \method{} learns meaningful control over latent computation from two perspectives: whether the router allocates latent operators according to task difficulty and reasoning stage, and whether the typed operators induce differentiated latent computation. An efficiency analysis is provided in Appendix~\ref{app:efficiency-analysis}.

\paragraph{Operator Specialization.}
We next study whether the three typed operators induce distinct functional roles. As shown in Figure~\ref{fig:pca-typed}(a), SoftCoT~\citep{xu2025softcot} mainly occupies a mixed latent region, whereas the representations conditioned on $O_g$, $O_s$, and $O_p$ form operator-aligned clusters in the same PCA space. This separation suggests that the operators do not collapse into homogeneous latent representations. The removal analysis in Figure~\ref{fig:pca-typed}(b) further supports this interpretation: removing $O_g$ leads to more strategy and setup errors, removing $O_s$ increases state and arithmetic errors, and removing $O_p$ produces more procedural and symbolic errors. Together, these results indicate that the typed operators learn complementary latent representations with distinct functional effects.

\paragraph{Task-Adaptive Invocation.}
Figure~\ref{fig:pca-typed}(c) reports the invocation ratios and relative positions of $O_s$ and $O_p$ on GSM8K and GPQA-Diamond. We visualize only $O_s$ and $O_p$, since $O_g$ is a global-orientation operator invoked at the beginning of generation. The router shows a clear task-adaptive pattern. On the relatively simple GSM8K task, the overall invocation ratio remains low, whereas GPQA-Diamond triggers latent operators more frequently, suggesting that the policy allocates more computation to harder reasoning tasks. The relative positions further reveal a stage-aware division of labor: $O_s$ is invoked more often in the early and middle stages of reasoning, where local states and intermediate conclusions are actively maintained; $O_p$ becomes more frequent in later stages, where reusable solution procedures are more likely to dominate. These results indicate that \method{} does not follow a rigid schedule, but learns to invoke latent computation according to both task complexity and reasoning state.

\paragraph{Token Budget Scaling.}
Figure~\ref{fig:latent-budget-scaling} studies the effect of the latent-call budget. 
Accuracy improves rapidly as $B_O$ increases from 0 to 5, showing that \method{} benefits from additional latent computation. 
However, when the budget increases further to $B_O=8$, the marginal accuracy gain becomes very small, while latency increases noticeably. 
Thus, $B_O=5$ provides a better accuracy--latency trade-off than using a larger budget.
\section{Conclusion}

We introduced \method{}, a framework that reframes latent reasoning as online, typed, and budgeted operator invocation.
By combining three typed latent operators with a budget-aware policy, \method{} enables autoregressive decoding to decide when to invoke internal computation and which latent operation to perform under a computation budget.
Across extensive experiments on three backbone LLMs, \method{} improves average performance by up to 14.49 points over CoT and by up to 4.30 points over the strongest competing baseline. 
It also generalizes across diverse reasoning domains and achieves the best final-stage performance with the lowest forgetting in sequential adaptation.
\section{Limitations}
\label{sec:limitations}

We note four limitations. First, our evaluation spans 1.5B--4B backbones across three model families; whether typed latent operators retain their advantage at substantially larger scales remains an open empirical question. Second, \method{} relies on rationale-like supervision to construct candidate boundaries in Stage~1. Although Stage~2 learns a routing policy over these candidates, fully unsupervised boundary discovery remains future work. Third, operator invocation triggers silent latent-token synthesis, which improves efficiency but reduces direct auditability compared with visible chain-of-thought rationales. The intended roles of the latent operators are learned rather than guaranteed by construction, so deployments should log operator identities, invocation positions, and budgets when traceability matters. Fourth, our experiments focus on reasoning benchmarks with verifier-style evaluation; broader open-ended generation tasks may require different rewards and additional safeguards against unfaithful or hard-to-inspect latent computation.

\section{Ethical Considerations}

This work studies latent reasoning as a mechanism for improving the efficiency and effectiveness of language-model reasoning. The experiments are conducted on public reasoning benchmarks and do not involve human subjects, private user data, or the collection of personally identifiable information. 

The main potential risk is dual use: improving the reasoning capability or inference efficiency of language models may also reduce the cost of deploying models in harmful applications. In addition, latent computation may make intermediate reasoning less transparent than explicit CoT generation, which could complicate debugging and auditing. Our work is intended as a research contribution on controllable latent computation rather than a deployed system. We encourage future applications of this method to include standard safety evaluations, misuse analysis, and appropriate monitoring when used in real-world settings.

\bibliography{custom}

\appendix
\section{Artifact}

\subsection{Artifact Licenses and Terms}

We use publicly available datasets, benchmarks, open-weight models, and evaluation tools for research purposes only. We cite the original creators of all artifacts and follow their stated licenses or terms of use. Our use of these artifacts is consistent with their intended research use: benchmark datasets are used for training, validation, or evaluation as appropriate, and held-out benchmark splits are used only for evaluation. We do not redistribute benchmark data or model weights as part of this submission.

Any released artifacts will be limited to code, configuration files, and reproduction scripts. We will not release derivative datasets or redistribute model weights. These artifacts are intended only for research reproducibility and further academic study.

\subsection{Artifact Documentation}

Table~\ref{tab:artifact-documentation} summarizes the main artifacts used in this work. 
All datasets and benchmarks are publicly available and are used for research purposes only. 
The language data used in our experiments is primarily English. 
The artifacts cover mathematical reasoning, scientific question answering, theorem-style reasoning, and code-generation tasks. 
They are not demographic or user-generated social datasets, and they are not used to study demographic groups, dialectal variation, or sociolinguistic phenomena. 
To the best of our knowledge, the benchmarks do not provide systematic demographic annotations for authors or represented populations; therefore, our analysis does not make demographic claims.

\begin{table*}[t]
\centering
\small
\begin{tabular}{llll}
\toprule
\textbf{Artifact} & \textbf{Type} & \textbf{Domain} & \textbf{Language / Format} \\
\midrule
GSM8K & Dataset & Grade-school math reasoning & English text \\
MATH-500 & Benchmark & Competition-level mathematics & English text / equations \\
GPQA-Diamond & Benchmark & Graduate-level science QA & English text \\
TheoremQA & Benchmark & Theorem-style reasoning & English text / equations \\
HumanEval & Benchmark & Code-generation evaluation & Python code and English prompts \\
SmolLM3-3B & Model & General-purpose language modeling & Text model \\
Qwen3-4B & Model & General-purpose language modeling & Text model \\
\bottomrule
\end{tabular}
\caption{Documentation of the main artifacts used in this work, including their type, domain coverage, and language or format.}
\label{tab:artifact-documentation}
\end{table*}

\subsection{Dataset Statistics}

Table~\ref{tab:dataset-statistics} reports the statistics of the datasets and benchmarks used in this work. 
We use the official splits whenever available. 
For benchmarks without a training split in our experiments, we use them only for evaluation. 
We do not collect new natural-language data. 
Synthetic reasoning traces or latent-operator training instances are generated only from the corresponding public training data and are used for research purposes.

\begin{table*}[t]
\centering
\small
\begin{tabular}{lccc}
\toprule
\textbf{Dataset / Benchmark} & \textbf{Train} & \textbf{Dev / Validation} & \textbf{Test / Eval} \\
\midrule
GSM8K & 7,473 & -- & 1,319 \\
MATH-500 & -- & -- & 500 \\
GPQA-Diamond & -- & -- & 198 \\
TheoremQA & -- & -- & 737 \\
HumanEval & -- & -- & 164 \\
\bottomrule
\end{tabular}
\caption{Statistics of the main datasets and benchmarks used in this work. Dashes indicate that the corresponding split is not used in our experiments.}
\label{tab:dataset-statistics}
\end{table*}
\section{Data Privacy and Content Safety}

We do not collect new user data or use private, personally identifying, or sensitive information. All datasets used in this work are publicly available reasoning, science, theorem-proving, and code-generation benchmarks. These benchmarks are used only for research purposes, following their intended use and access conditions.

To check for privacy and content risks, we reviewed the dataset descriptions and manually inspected representative examples from the training, validation, and evaluation data used in our experiments. We found no fields designed to identify individual people, such as names linked to real individuals, addresses, phone numbers, email addresses, account identifiers, or other personally identifying information. The benchmarks also do not target offensive, hateful, or abusive language generation. Therefore, no additional anonymization was required. We do not redistribute the benchmark data or model weights as part of this submission.
\section{Use of AI Assistants}

We used AI writing assistance tools only for language polishing, grammar checking, and improving the clarity of author-written text. These tools were not used to generate research ideas, design the method, conduct experiments, analyze results, or produce conclusions. All AI-assisted edits were reviewed, verified, and revised by the authors, who take full responsibility for the content of the paper.
\section{Additional Related Works}
\label{app:additional-related-works}

\paragraph{Continual learning.}
Continual learning studies sequential adaptation under the risk of catastrophic interference, where learning a new task degrades performance on earlier tasks~\citep{mccloskey1989catastrophic,delange2021continual}. Representative strategies include regularizing important parameters to preserve previous behavior~\citep{kirkpatrick2017overcoming}, distilling old-task predictions during new-task training~\citep{li2018learning}, replaying or generating samples from past tasks in lifelong language learning~\citep{sun2019lamol}, and allocating lightweight prompt parameters across a task stream~\citep{razdaibiedina2023progressive}. These methods primarily protect previously acquired behavior by constraining parameter updates, revisiting past data, or assigning task-specific capacity. Our evaluation is complementary: it does not introduce a general continual-learning algorithm, but asks whether typed latent operators can serve as a low-interference adaptation interface during sequential reasoning-task updates.

\paragraph{Multi-task learning.}
Multi-task learning improves transfer by training a shared model over multiple related objectives~\citep{caruana1997multitask,ruder2017overview}. In NLP, this idea has appeared in shared neural architectures for multiple language-processing tasks~\citep{collobert2008unified} and, more recently, in large-scale prompted task mixtures that induce zero-shot generalization~\citep{sanh2022multitask}. Unlike multi-task training, the continual-learning protocol in Appendix~\ref{app:continual-learning} exposes tasks sequentially rather than as a joint mixture. \method{} is orthogonal to both settings: instead of relying solely on a monolithic shared representation, it provides a small set of typed latent computation paths and learns when to route a reasoning trajectory through them. This makes the experiment a test of whether operator-mediated latent computation reduces interference while preserving cross-task transfer.

\section{Experiment Details}
\label{app:experiment-details}

\subsection{Training Data}
\label{app:training-data}

We use 15K problem--solution traces sampled from OpenR1-Math~\citep{openr1} to train \method{}. The same training pool is used for Stage~1 operator-conditioned latent-token synthesis and Stage~2 budgeted routing, while the backbone-specific SFT and GRPO baselines use the same data and prompt templates for controlled comparison. Evaluation benchmarks are used only for testing, and no test split is used for model selection.

\subsection{Benchmarks and Metrics}
\label{app:benchmark-details}

We evaluate four reasoning benchmarks. GSM8K~\citep{cobbe2021gsm8k} and MATH-500~\citep{hendrycks2021math} evaluate mathematical reasoning at different levels of symbolic complexity. GPQA-Diamond~\citep{rein2023gpqa} evaluates graduate-level scientific reasoning. TheoremQA~\citep{chen2023theoremqa} evaluates theorem-oriented reasoning that requires applying formal concepts and structured solution patterns.

All main results report Pass@1 accuracy under greedy decoding. The macro-average is computed as the unweighted average over GSM8K, MATH-500, GPQA-Diamond, and TheoremQA.
For all benchmarks, we require the model to wrap its final answer with \verb|\boxed{}|. Outputs without a boxed final answer are judged as incorrect. For boxed outputs, we extract the content inside \verb|\boxed{}| and apply the benchmark-specific answer normalizer or verifier.

\subsection{Baselines and Budget Matching}
\label{app:baseline-details}

We compare against explicit-reasoning baselines, including CoT~\citep{wei2022chain}, SFT~\citep{ouyang2022training}, and GRPO~\citep{shao2024deepseekmath}, and latent-reasoning baselines, including SoftCoT~\citep{xu2025softcot}, Soft-Thinking~\citep{zhang2025softthinking}, MemGen~\citep{zhang2025memgen}, and SwiReasoning~\citep{shi2025swireasoning}. All baselines use the same backbone checkpoint, tokenizer, prompt template, and decoding setting whenever applicable. 
Baseline-specific hyperparameters follow the authors' recommended settings when available; otherwise, we tune them on the same development split used for \method{}.

\subsection{Implementation Details}
\label{app:implementation-details}

Table~\ref{tab:implementation-details} summarizes the implementation configuration.

\begin{table*}[t]
\centering
\small
\setlength{\tabcolsep}{4pt}
\begin{tabular}{L{0.20\linewidth}L{0.25\linewidth}L{0.22\linewidth}L{0.22\linewidth}}
\toprule
\textbf{Configuration} & \textbf{Parameter} & \textbf{Stage 1} & \textbf{Stage 2} \\
\midrule
\multirow{6}{*}{Model}
& Backbones & \multicolumn{2}{L{0.48\linewidth}}{Qwen2.5-1.5B-Instruct, SmolLM3-3B, Qwen3-4B} \\
& PEFT Method & LoRA & LoRA \\
& LoRA rank & 8 & 8 \\
& LoRA alpha & 16 & 16 \\
& LoRA dropout & 0.1 & 0.1 \\
& Target modules & \texttt{q\_proj}, \texttt{v\_proj} & \texttt{q\_proj}, \texttt{v\_proj} \\
\midrule
\multirow{5}{*}{Latent operators}
& Operator types & \multicolumn{2}{L{0.48\linewidth}}{$O_g$, $O_s$, $O_p$} \\
& Synthesizer LoRA Parameters & Random initialization & Frozen from Stage 1 \\
& Latent length $N_g$ & 8 & 8 \\
& Latent length $N_s$ & 4 & 4 \\
& Latent length $N_p$ & 4 & 4 \\
\midrule
\multirow{8}{*}{Training}
& Training data & \multicolumn{2}{L{0.48\linewidth}}{15K OpenR1-Math traces} \\
& Batch size & 8 & 8 \\
& Epochs / steps & 2 epochs & 1 epochs \\
& Learning rate & $1\times10^{-5}$ & $1\times10^{-5}$ \\
& Optimizer & AdamW & AdamW \\
& Scheduler & Cosine & Cosine \\
& Warmup ratio & 0.1 & 0.1 \\
& Random seeds & 42 \\
\midrule
\multirow{7}{*}{GRPO-Relative}
& GRPO group size $G$ & -- & 8 \\
& Clip ratio $\epsilon$ & -- & 0.2 \\
& KL coefficient $\beta$ & -- & 0.03 \\
& Latent-call budget $B_O$ & -- & 5 \\
& Budget penalty $\lambda$ & -- & 0.1 \\
& Anchor weight $\alpha$ & -- & 0.1 \\
& Success threshold $\rho_0$ & -- & verifier success \\
\midrule
\multirow{4}{*}{Evaluation}
& Decoding & \multicolumn{2}{L{0.48\linewidth}}{Greedy decoding} \\
& Max new tokens & \multicolumn{2}{L{0.48\linewidth}}{2048} \\
& Answer extractor & \multicolumn{2}{L{0.48\linewidth}}{\texttt{\textbackslash boxed\{\}} parser with benchmark-specific normalization} \\
\bottomrule
\end{tabular}
\caption{Implementation configuration for \method{}. Stage~2 uses a per-trajectory latent-call budget of $B_O=5$; task success is the benchmark-specific exact-match or verifier score.}
\label{tab:implementation-details}
\end{table*}

\subsection{Result Reporting.}
Unless otherwise specified, all results are reported as pass@1 accuracy on the official evaluation split of each benchmark. 
Each score is computed over all examples in the corresponding evaluation set, and the average score is the arithmetic mean across benchmarks. 
We report single-run results rather than the maximum over multiple random seeds or prompt trials. 
For training-based methods, we use a fixed random seed and keep the decoding and evaluation protocol identical across methods. 
We do not select the best result from repeated runs.
\section{Training Details}
\label{app:training-details}

We train \method{} in two stages. Stage~1 learns typed latent operators while keeping the backbone LLM frozen. Stage~2 freezes the learned synthesizer and operator-specific components, and trains the backbone to decide \emph{when} to invoke latent computation and \emph{which} operator to select, under an explicit budget.
Implementation hyperparameters are summarized in Appendix~\ref{app:implementation-details}.

\subsection{Stage 1: Latent Synthesis Optimization}
\label{app:stage1}

In Stage~1, the backbone LLM with parameters $\theta$ is frozen, and we optimize the latent synthesizer $\phi$ together with the operator-specific components $\{Q_k,\mathrm{Proj}_k\}_{O_k\in\mathcal{O}}$. For brevity, we collect all Stage~1 trainable parameters as $\Phi = (\phi, \{Q_k, \mathrm{Proj}_k\}_{O_k \in \mathcal{O}})$. For each supervised instance $(q,y_{1:T})$, we construct a set of candidate boundary--operator pairs $\mathcal{B}(q,y)$ by parsing the target sequence with structure-aware rules:

\begin{itemize}[leftmargin=*,labelsep=0.5em,topsep=2pt,itemsep=1pt]
  \item $O_g$ is associated with \emph{answer starts}: the first position immediately following the question, and any position following solution-onset cues such as ``Let me solve'', ``Solution:'', or the analogous markers in code/math templates.
  \item $O_s$ is associated with \emph{reasoning-step boundaries}: positions following \verb|\n\n| or enumerated step markers (``Step $i$:'', ``$i$.\,'') detected by a lightweight regex parser.
  \item $O_p$ is associated with \emph{structural boundaries}: positions immediately before formula spans (\verb|$...$|), fenced code blocks (\verb|```|), or structured output schemata (e.g., JSON keys).
\end{itemize}

For each training step, we sample a pair $(b,k)\sim\mathcal{B}(q,y)$ and construct the prefix embedding sequence $x_b$ from the question and the ground-truth prefix $y_{<b}$. The selected operator produces latent tokens $z_b^k = O_k(x_b)$ via Eqs.~\eqref{eq:synth-intermediate}--\eqref{eq:synth-final}, which are then appended to $x_b$ following Eq.~\eqref{eq:lcf-transition}. We do not directly supervise $z_b^k$; instead, the latent tokens are trained only through their contribution to predicting subsequent visible tokens. Let $\tilde{c}_t$ denote the latent-augmented context preceding $y_t$ (i.e., $\tilde{c}_t = (x_b, z_b^k, y_{b:t-1})$ for $t \ge b$, and the standard prefix otherwise). The sampled boundary only determines \emph{where} latent tokens are appended; the loss is applied to the \emph{entire} target sequence so that gradients flow back through $z_b^k$ to $\Phi$ from every downstream position, preventing the latent module from overfitting to local suffix prediction:
\begin{equation}
\mathcal{L}_{\mathrm{stage1}}
= -\sum_{t=1}^{T}\log p_{\theta,\,\Phi}\!\left(y_t \mid \tilde{c}_t\right),
\label{eq:app-operator-sft}
\end{equation}
where the gradient updates only $\Phi$ since $\theta$ is frozen.

\subsection{Stage 2: Operator Invocation Optimization}
\label{app:stage2}

Stage~2 learns a step-level operator invocation policy that decides whether to emit a visible token or to invoke a typed latent operator. We freeze $\Phi$ from Stage~1 and optimize the backbone parameters $\theta$ together with the extended head rows $\psi$ introduced in §\ref{sec:method-operators}.

\paragraph{Group-relative policy optimization.}
For each prompt $q$, the current policy samples a group of trajectories $\mathcal{G}(q)=\{\tau_i\}_{i=1}^{G}$. Each trajectory receives the same budget-aware reward used in the main text:
\begin{equation}
R(\tau)=R_{\mathrm{task}}(\tau)-\lambda\,P_{\mathrm{bud}}(\tau),
\label{eq:app-trajectory-reward}
\end{equation}
where
\begin{equation}
P_{\mathrm{bud}}(\tau)=s(\tau)(n_{\mathrm{op}}(\tau)-B_O)_+ .
\label{eq:app-budget-penalty}
\end{equation}
Here, $s(\tau)\in\{0,1\}$ indicates task success, $n_{\mathrm{op}}(\tau)$ is the number of latent-computation calls in $\tau$, and $B_O$ is the per-trajectory call budget. The penalty is non-zero only on \emph{successful, over-budget} trajectories: failures contribute no budget penalty so that exploration on hard instances is not suppressed, and under-budget successes are free so that the model has no incentive to remove useful latent computation.

Let $\bar R_q$ and $\sigma_q$ denote the mean and standard deviation of rewards in $\mathcal{G}(q)$. GRPO assigns each trajectory a group-relative advantage
\begin{equation}
A_i=\frac{R(\tau_i)-\bar R_q}{\sigma_q+\epsilon}.
\label{eq:app-grpo-advantage}
\end{equation}
For any visible-token or operator-invocation action $a_{i,j}$ at context $x_{i,j}$, define the importance-sampling ratio as
\begin{equation}
r_{i,j}
=
\frac{p_{\theta,\psi}(a_{i,j}\mid x_{i,j})}
{p_{\mathrm{old}}(a_{i,j}\mid x_{i,j})}.
\label{eq:app-grpo-ratio}
\end{equation}
Let $\bar r_{i,j}=\mathrm{clip}(r_{i,j},1-\epsilon,1+\epsilon)$ denote the clipped ratio. The clipped per-action surrogate is
\begin{equation}
\ell_{i,j}^{\mathrm{clip}}
=
\min\!\left(r_{i,j}A_i,\,\bar r_{i,j}A_i\right).
\label{eq:app-grpo-token}
\end{equation}
The clipped GRPO objective~\citep{shao2024deepseekmath,guo2025deepseek} is
\begin{equation}
\begin{aligned}
\mathcal{J}_{\mathrm{GRPO}}(R)
&=
\frac{1}{G}\sum_{i=1}^{G}\frac{1}{|\tau_i|}
\sum_{j\in\tau_i}
\ell_{i,j}^{\mathrm{clip}}\\
&\quad
-\beta\,D_{\mathrm{KL}}\!\left(p_{\theta,\psi}\,\|\,p_{\mathrm{ref}}\right),
\end{aligned}
\label{eq:app-grpo-objective}
\end{equation}
and the minimized GRPO loss is $\mathcal{L}_{\mathrm{GRPO}}(R)=-\mathcal{J}_{\mathrm{GRPO}}(R)$.

\paragraph{Operator-anchored supervision.}
The GRPO loss applies the same trajectory-level advantage across all actions in a sampled response. This is sufficient for task-level learning, but operator invocations are sparse relative to visible-token actions, so their selection gradients can be diluted. We therefore add an operator-anchored auxiliary objective that applies the same group-relative advantage directly to operator invocations.

For each sampled trajectory $\tau_i$, let
\begin{equation}
M_{\mathcal{O}}(\tau_i)=\{\,j\mid a_{i,j}\in\mathcal{O}\,\}
\label{eq:app-operator-mask}
\end{equation}
be the set of positions where the policy invokes a latent operator, and let $N_{\mathcal{O}}=\sum_i |M_{\mathcal{O}}(\tau_i)|$. The auxiliary loss is
\begin{equation}
\begin{aligned}
\mathcal{L}_{\mathrm{anch}}
&=
-\frac{1}{N_{\mathcal{O}}}
\sum_{i=1}^{G}\sum_{j\in M_{\mathcal{O}}(\tau_i)}
\ell_{i,j}^{\mathrm{clip}}.
\end{aligned}
\label{eq:app-operator-anchor-loss}
\end{equation}
When $N_{\mathcal{O}}=0$, we set $\mathcal{L}_{\mathrm{anch}}=0$. This term reuses the GRPO advantage rather than introducing extra intervention rollouts, and concentrates additional gradient on the operator invocations that trigger typed latent operators.

\paragraph{Combined objective.}
The Stage~2 loss combines the standard GRPO loss with the operator-anchored auxiliary term:
\begin{equation}
\min_{\theta,\psi}\ 
\mathcal{L}_{\mathrm{stage2}}
=
\mathcal{L}_{\mathrm{GRPO}}(R)
+\alpha\,\mathcal{L}_{\mathrm{anch}},
\label{eq:app-stage2-control-loss}
\end{equation}
where $\alpha$ controls the auxiliary weight. We tune $(\lambda, B_O, \alpha, G)$ on the development split.

\begin{table*}[!t]
  \centering
  \setlength{\tabcolsep}{4pt}
  \begin{tabular}{lccccccc}
    \toprule
    \textbf{Method} & \textbf{GSM8K} & \textbf{MATH-500} & \textbf{GPQA-Diamond} & \textbf{TheoremQA} & \textbf{Average} & $\Delta_{\mathrm{CoT}}$ \\
    \midrule
    CoT                 & 64.67 & 52.80 & 17.17 & 18.07 & 38.17 & -- \\
    SFT                 & 66.49 & 47.20 & 19.70 & 15.93 & 37.33 & $-0.84$ \\
    GRPO                & \underline{74.30} & \underline{54.80} & 12.12 & \underline{19.41} & 40.16 & $+1.99$ \\
    \midrule
    SoftCoT             & 51.55 & 41.80 & 19.19 & 13.92 & 31.62 & $-6.55$ \\
    Soft-Thinking       & 68.92 & 51.60 & 11.11 & 16.87 & 37.13 & $-1.04$ \\
    MemGen              & 71.19 & 53.20 & \underline{21.21} & 18.62 & 41.06 & $+2.89$ \\
    SwiReasoning        & 61.71 & 43.60 & 11.11 & 17.62 & 33.51 & $-4.66$ \\
    \midrule
    \rowcolor{cvprblue!16}
    \textbf{\method{} (Ours)} & & & & & & \\
    \quad + Stage 1     & 73.39 & 53.00 & \underline{21.21} & 19.01 & \underline{41.65} & $\underline{+3.48}$ \\
    \quad + Stage 2     & \textbf{75.28} & \textbf{57.60} & \textbf{23.23} & \textbf{20.08} & \textbf{44.05} & $\textbf{+5.88}$ \\
    \bottomrule
  \end{tabular}
  \caption{Pass@1 (\%, greedy decoding) on Qwen2.5-1.5B-Instruct. \textbf{Best} and \underline{second-best} among complete rows; $\Delta_{\mathrm{CoT}}$ is the macro-average gain over CoT. All latent baselines budget-matched to within $\pm10\%$ of \method{}.}
  \label{tab:qwen25-main}
\end{table*}

\section{Results on Qwen2.5-1.5B-Instruct}
\label{app:qwen25-results}

We report the full evaluation on Qwen2.5-1.5B-Instruct \citep{yang2024qwen25} in this appendix because the 1.5B regime exposes several small-model failure modes for prior latent-reasoning methods that are absent or weaker on the SmolLM3-3B/Qwen3-4B backbones used in the main paper. Setup, baselines, and metrics follow §\ref{sec:exp-setup}; all latent baselines are budget-matched to within $\pm10\%$ of \method{}'s total visible + latent token count, and all numbers are Pass@1 under greedy decoding.

\paragraph{Small-model regime exposes latent-baseline brittleness.}
At 1.5B, three patterns emerge that motivate the design of \method{}. First, the \emph{implicit} latent baselines do not transfer down: SoftCoT collapses to $31.62$ on the macro-average ($-6.55$ vs.~CoT), and Soft-Thinking matches CoT on average but loses $6.06$ points on GPQA-Diamond. This is consistent with the original SoftCoT report \citep{xu2025softcot} that the assistant-generated soft thoughts require a sufficiently expressive backbone to be consumed productively. Second, the \emph{hybrid} baselines are uneven: MemGen becomes the strongest prior latent baseline with a $+2.89$ macro-average gain over CoT, but entropy-based switching with SwiReasoning still falls below CoT. Third, RL alone (GRPO, $+1.99$ over CoT) wins GSM8K, MATH-500, and TheoremQA among prior baselines, but its GPQA-Diamond accuracy ($12.12$) regresses below CoT, suggesting that the policy improvements concentrate on the mathematical subset on which the reward is trained.

\paragraph{\method{} dominates the macro-average and the knowledge-intensive split.}
\method{} (Stage 1) attains a strong macro-average ($41.65$), exceeding the strongest prior baseline MemGen by $+0.59$ points and CoT by $+3.48$. On GPQA-Diamond, \method{} reaches $21.21$, matching MemGen and improving over GRPO by $+9.09$ points. On the remaining three benchmarks \method{} is within $0.40$--$1.80$ points of the best prior baseline. The Stage~2 row further raises the average to $44.05$ after adding budget-aware routing and operator-anchored supervision.

\begin{figure*}[t]
\centering
\includegraphics[width=\textwidth]{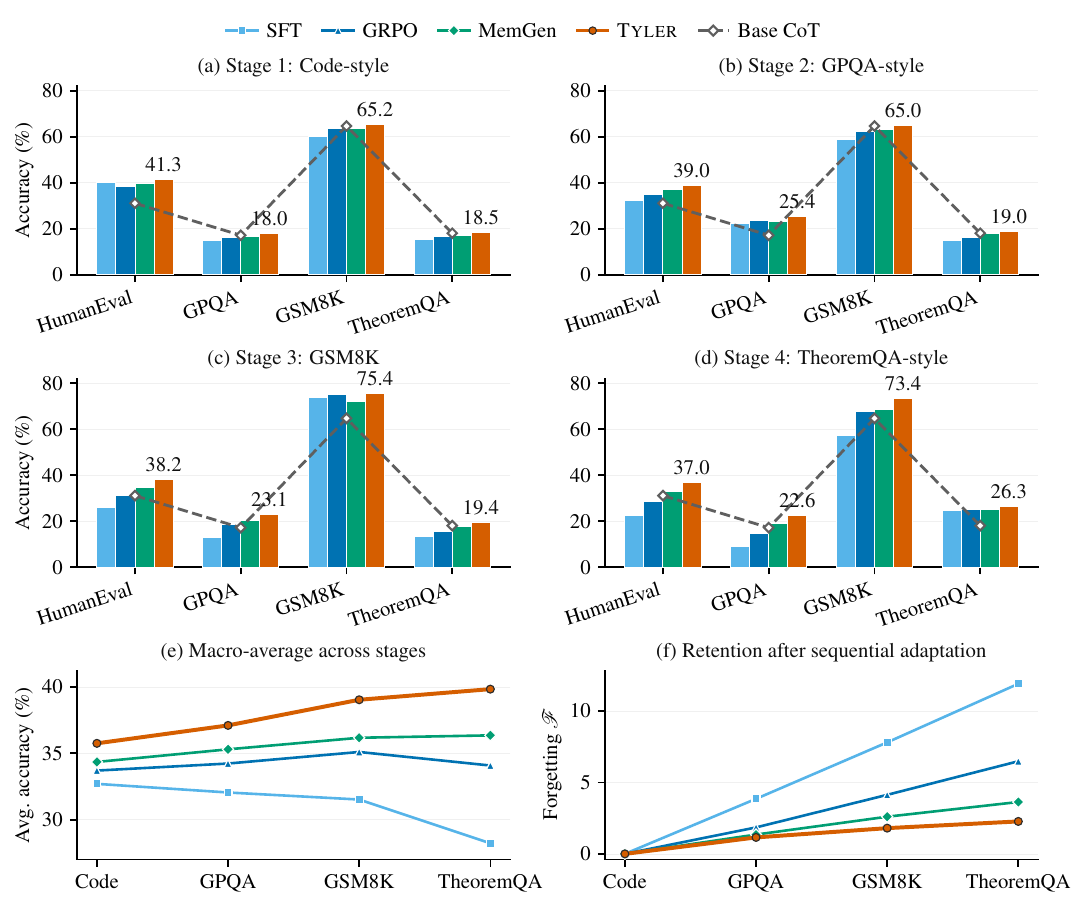}
\caption{Fixed-order continual-learning evaluation on Qwen2.5-1.5B-Instruct. Panels (a)--(d) show benchmark performance after each sequential adaptation stage, with tasks on the horizontal axis and methods grouped by color. The dashed gray line denotes the base Vanilla CoT model. Panels (e)--(f) summarize the macro-average and the stage-wise forgetting score $\mathcal{F}$. Lower $\mathcal{F}$ indicates stronger retention of previously observed task capabilities.}
\label{fig:continual-learning}
\end{figure*}

\section{Continual-Learning Evaluation}
\label{app:continual-learning}

This appendix presents a fixed-order continual-learning evaluation to examine whether synthetic latent tokens provide a low-interference interface for sequential adaptation. We use Qwen2.5-1.5B-Instruct as the backbone and adapt each method sequentially across four task families: code-style reasoning evaluated on HumanEval~\citep{chen2021humaneval} $\rightarrow$ GPQA-style science reasoning $\rightarrow$ GSM8K arithmetic reasoning $\rightarrow$ TheoremQA-style theorem reasoning. After each adaptation stage, the resulting model is evaluated on all four benchmarks, including both the newly adapted domain and previously observed domains. For each stage, adaptation uses training or development traces from the corresponding task family, while the held-out benchmark splits are used exclusively for evaluation. This setup provides a controlled evaluation of sequential interference under a fixed task order, rather than an exhaustive continual-learning benchmark over all possible task permutations.

We report the macro-average over the four evaluation benchmarks and the stage-wise forgetting score:
\begin{equation}
\mathcal{F}_t =
\frac{1}{|\mathcal{S}_t|}
\sum_{i\in\mathcal{S}_t}
\left(
\max_{s\leq t} A_{s,i} - A_{t,i}
\right),
\end{equation}
where $\mathcal{S}_t$ denotes the set of task families observed up to stage $t$, and $A_{t,i}$ is the accuracy on task family $i$ after the $t$-th adaptation stage. Lower values of $\mathcal{F}_t$ indicate stronger retention of previously acquired capabilities. Since the newly introduced task at a given stage contributes zero forgetting by construction, the forgetting score at stage 1 is trivially zero for all methods.

\begin{table*}[t]
\centering
\setlength{\tabcolsep}{3pt}
\begin{tabular}{cllcccccc}
\toprule
\textbf{Stage}$\downarrow$ & \textbf{Adapted Domain} & \textbf{Method} & \textbf{HumanEval} & \textbf{GPQA} & \textbf{GSM8K} & \textbf{TheoremQA} & \textbf{Avg.} & $\mathcal{F}\downarrow$ \\
\midrule
-- & Base & Vanilla CoT & 31.10 & 17.17 & 64.67 & 18.07 & 32.75 & -- \\
\midrule
\multirow{4}{*}{1}
& \multirow{4}{*}{HumanEval}
& SFT & 40.20 & 14.90 & 60.30 & 15.40 & 32.70 & 0.00 \\
& & GRPO & 38.50 & 16.10 & 63.40 & 16.80 & 33.70 & 0.00 \\
& & MemGen & 39.60 & 16.70 & 63.80 & 17.30 & 34.35 & 0.00 \\
& & \method{} (Ours) & 41.30 & 18.00 & 65.20 & 18.50 & 35.75 & 0.00 \\
\midrule
\multirow{4}{*}{2}
& \multirow{4}{*}{GPQA}
& SFT & 32.50 & 22.20 & 58.70 & 14.80 & 32.05 & 3.85 \\
& & GRPO & 34.80 & 23.70 & 62.20 & 16.20 & 34.23 & 1.85 \\
& & MemGen & 36.90 & 23.20 & 63.30 & 17.80 & 35.30 & 1.35 \\
& & \method{} (Ours) & 39.00 & 25.40 & 65.00 & 19.00 & 37.10 & 1.15 \\
\midrule
\multirow{4}{*}{3}
& \multirow{4}{*}{GSM8K}
& SFT & 26.10 & 12.90 & 73.60 & 13.50 & 31.52 & 7.80 \\
& & GRPO & 31.20 & 18.60 & 74.90 & 15.70 & 35.10 & 4.13 \\
& & MemGen & 34.80 & 20.20 & 72.20 & 17.50 & 36.17 & 2.60 \\
& & \method{} (Ours) & 38.20 & 23.10 & 75.40 & 19.40 & 39.03 & 1.80 \\
\midrule
\multirow{4}{*}{4}
& \multirow{4}{*}{TheoremQA}
& SFT & 22.40 & 8.80 & 57.20 & 24.50 & 28.23 & 11.90 \\
& & GRPO & 28.60 & 14.80 & 67.80 & 25.10 & 34.08 & 6.48 \\
& & MemGen & 32.90 & 19.00 & 68.60 & 24.90 & 36.35 & 3.63 \\
\rowcolor{cvprblue!16}
& & \method{} (Ours) & 37.00 & 22.60 & 73.40 & 26.30 & 39.83 & 2.27 \\
\bottomrule
\end{tabular}
\caption{Fixed-order continual-learning evaluation on Qwen2.5-1.5B-Instruct. The downward arrow in the Stage column indicates the top-to-bottom sequential adaptation order. The macro-average is computed over all four evaluation benchmarks at every stage. The forgetting score $\mathcal{F}$ is computed over task families observed up to the current stage, with lower values indicating better retention.}
\label{tab:continual-learning}
\end{table*}

\paragraph{Improved Adaptation under Sequential Training.}
Across the four-stage adaptation sequence, \method{} achieves the highest macro-average after every stage. Its advantage over the strongest baseline increases from $+1.40$ points after code-style adaptation to $+3.48$ points after the final TheoremQA-style adaptation, where \method{} reaches $39.83$ compared with $36.35$ for MemGen, $34.08$ for GRPO, and $28.23$ for SFT. Under this fixed task order, the performance gap becomes larger as more adaptation stages are applied, suggesting that \method{} maintains stronger overall performance during sequential adaptation.

\paragraph{Reduced Forgetting on Previously Observed Tasks.}
\method{} also obtains the lowest forgetting score throughout the sequence. At stage 2, it achieves $\mathcal{F}=1.15$, slightly below MemGen at $1.35$ and below GRPO and SFT at $1.85$ and $3.85$, respectively. The gap becomes more pronounced in later stages. After the final adaptation stage, \method{} obtains $\mathcal{F}=2.27$, reducing forgetting by $1.36$ points relative to MemGen, $4.21$ points relative to GRPO, and $9.63$ points relative to SFT. These results suggest that operator-based latent adaptation can reduce interference with previously acquired capabilities compared with direct SFT or policy optimization alone.

\paragraph{A Better Stability--Plasticity Trade-Off.}
After the final TheoremQA-style adaptation, \method{} achieves the best score on all four benchmarks: $37.00$ on HumanEval, $22.60$ on GPQA, $73.40$ on GSM8K, and $26.30$ on TheoremQA. Compared with MemGen, the gains are $+4.10$, $+3.60$, $+4.80$, and $+1.40$ points, respectively. SFT shows substantial improvement on the currently adapted TheoremQA-style task but suffers severe degradation on previously adapted domains, leading to the highest final forgetting score of $11.90$. GRPO and MemGen mitigate this degradation to some extent, while \method{} achieves both the highest final macro-average and the lowest final forgetting score. This indicates a stronger stability--plasticity trade-off in the tested sequential adaptation setting.
\section{Computational Efficiency Analysis}
\label{app:efficiency-analysis}

\subsection{Experimental Settings}
\label{app:efficiency-protocol}

We report end-to-end per-example inference time together with task performance. All methods are evaluated with greedy decoding, batch size 1, identical prompt templates, the same maximum generation length, and the same Qwen3-4B backbone. Latency is measured after tokenization and excludes data loading, answer extraction, verifier execution, and logging.
For each method, we record visible generated tokens and latent tokens. The total token budget is the sum of visible and latent tokens, so latent computation is accounted for explicitly rather than treated as free computation.

\subsection{Accuracy--Latency Tradeoff}
\label{app:efficiency-summary}

\begin{figure*}[t]
  \centering
  \includegraphics[width=\textwidth]{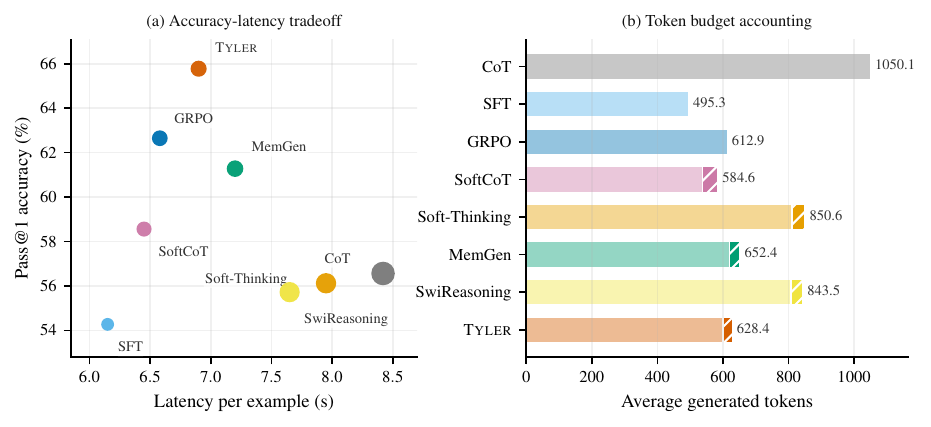}
  \caption{Computational efficiency on Qwen3-4B, averaged across GSM8K, MATH-500, TheoremQA, and GPQA-Diamond. Each method uses a fixed color across panels. In the token accounting panel, lighter bar segments denote visible tokens and hatched saturated caps denote latent tokens.}
  \label{fig:efficiency-summary}
\end{figure*}

Figure~\ref{fig:efficiency-summary} summarizes the accuracy--efficiency tradeoff across GSM8K, MATH-500, TheoremQA, and GPQA-Diamond. The left panel reports macro-average Pass@1 against per-example latency, with marker area proportional to the total generated-token budget. The right panel decomposes the same budget into visible and latent tokens.

\method{} reaches the highest average accuracy while staying within a compact compute budget. Compared with CoT, \method{} improves average Pass@1 from $56.56$ to $65.78$ while reducing latency from $8.42$s to $6.90$s and reducing the total token budget from $1050.1$ to $628.4$ tokens. Compared with the strongest prior baseline by accuracy, GRPO, \method{} gains $+3.13$ points with a modest $0.32$s latency increase. Compared with MemGen, \method{} is both more accurate and faster: it improves average Pass@1 by $+4.50$ points, reduces latency by $0.30$s, and uses fewer total tokens ($628.4$ vs.\ $652.4$).

\subsection{Per-Benchmark Overhead}
\label{app:efficiency-overhead}

Table~\ref{tab:efficiency-overhead} breaks down \method{} latency by benchmark, ordered from easier to harder tasks. The total latency increases with task difficulty, from GSM8K to GPQA-Diamond, while the wall-clock time spent in latent synthesis remains a small fraction of the overall inference time.

\begin{table}[t]
\centering
\small
\setlength{\tabcolsep}{4pt}
\begin{tabular}{lccc}
\toprule
\textbf{Benchmark} & \textbf{Total (s)} & \textbf{Synth. (s)} & \textbf{Synth. Share (\%)} \\
\midrule
GSM8K & 4.90 & 0.38 & 7.8 \\
MATH & 6.30 & 0.49 & 7.8 \\
Theorem & 7.49 & 0.54 & 7.2 \\
GPQA & 8.90 & 0.58 & 6.5 \\
\bottomrule
\end{tabular}
\caption{Per-benchmark latency breakdown for \method{} on Qwen3-4B. \emph{Synth.} denotes wall-clock time spent in latent reasoning.}
\label{tab:efficiency-overhead}
\end{table}

The synthesis overhead remains stable across benchmarks: latent reasoning accounts for $6.5$--$7.8\%$ of total inference time. The larger latency on TheoremQA and GPQA-Diamond is therefore driven primarily by longer visible reasoning traces rather than an uncontrolled growth in the latent synthesis module. This behavior matches the intended design of \method{}: the router allocates a small number of typed latent computations while the base decoder remains responsible for producing the final visible answer.

\subsection{Budget Sweep}
\label{app:efficiency-budget}

Table~\ref{tab:efficiency-budget} varies the latent budget on Qwen3-4B. The sweep tests whether \method{} benefits from additional latent computation and whether the gains saturate after a small budget.

\begin{table}[t]
\centering
\small
\setlength{\tabcolsep}{5pt}
\begin{tabular}{cccc}
\toprule
$B_O$ & \textbf{Accuracy (\%)} & \textbf{Latency (s)} & \textbf{Latent Tokens} \\
\midrule
0 & 60.11 & 6.35 & 0 \\
1 & 62.44 & 6.62 & 7 \\
2 & 64.08 & 6.85 & 13 \\
5 & 65.78 & 6.90 & 28 \\
8 & 65.95 & 7.85 & 44 \\
\bottomrule
\end{tabular}
\caption{Budget sweep for \method{} on Qwen3-4B, averaged across GSM8K, MATH-500, TheoremQA, and GPQA-Diamond.}
\label{tab:efficiency-budget}
\end{table}

Most of the gain appears under a small latent budget. Moving from $B_O=0$ to $B_O=2$ improves accuracy by $+3.97$ points with only $0.50$s additional latency. The full budget $B_O=5$ further improves accuracy to $65.78$, yielding a $+5.67$ point gain over no latent routing. Increasing the budget to $B_O=8$ adds $16$ latent tokens and $0.95$s latency but improves accuracy by only $0.17$ points, indicating that the accuracy benefit saturates once the router has enough budget to invoke the most useful latent operators.

\end{document}